\documentclass[sigconf,natbib=false]{acmart}
\usepackage[utf8]{inputenc}
\usepackage{newunicodechar}
\newunicodechar{̈}{"{}}
\usepackage[T1]{fontenc}

\usepackage{placeins}
\usepackage{graphicx}
\usepackage{algorithm}
\usepackage{algorithmic}
\usepackage{booktabs}
\usepackage{multirow}
\usepackage{array}
\usepackage{svg}

\usepackage{colortbl}
\definecolor{lightgray}{gray}{0.9}
\newcommand{\mycc}{\cellcolor{lightgray}}

\AtBeginDocument{%
  }

\RequirePackage[
  datamodel=acmdatamodel,
  style=acmnumeric,
  ]{biblatex}

\begin{document}

\title{Addressing Antisocial Behavior in Multi-Party Dialogs Through Multimodal Representation Learning}
\renewcommand{\shorttitle}{Multimodal Representation Learning for Antisocial Behavior in Multi-Party Dialogs}

\author{Hajar Bakarou}
\email{hajar.bakarou@etu.univ-cotedazur.fr}
\affiliation{%
  \institution{Universit\'e C{\^{\o}}te d'Azur, CNRS, Inria, I3S}
  \city{Sophia Antipolis}
  \country{France}
}
\author{Mohamed Sinane El Messoussi}
\email{mohamed.elmessoussi@etu.univ-cotedazur.fr}
\affiliation{%
  \institution{Universit\'e C{\^{\o}}te d'Azur, CNRS, Inria, I3S}
  \city{Sophia Antipolis}
  \country{France}
}
\author{Ana\"is Ollagnier}
\email{ollagnier@i3s.unice.fr}
\affiliation{%
  \institution{Universit\'e C{\^{\o}}te d'Azur, CNRS, Inria, I3S}
  \city{Sophia Antipolis}
  \country{France}
}
\orcid{0000-0003-4117-1191}


\begin{abstract}
Antisocial behavior (ASB) on social media---including hate speech, harassment, and cyberbullying---poses growing risks to platform safety and societal well-being. Prior research has focused largely on networks such as X and Reddit, while \textit{multi-party conversational settings} remain underexplored due to limited data. To address this gap, we use \textit{CyberAgressionAdo-Large}, a French open-access dataset simulating ASB in multi-party conversations, and evaluate three tasks: \textit{abuse detection}, \textit{bullying behavior analysis}, and \textit{bullying peer-group identification}. We benchmark six text-based and eight graph-based \textit{representation-learning methods}, analyzing lexical cues, interactional dynamics, and their multimodal fusion. Results show that multimodal models outperform unimodal baselines. The late fusion model \texttt{mBERT + WD-SGCN} achieves the best overall results, with top performance on abuse detection (0.718) and competitive scores on peer-group identification (0.286) and bullying analysis (0.606). Error analysis highlights its effectiveness in handling nuanced ASB phenomena such as implicit aggression, role transitions, and context-dependent hostility.

\end{abstract}



\maketitle

\section{Introduction}

Social media platforms have transformed how people consume information, engage in discourse, and form communities~\cite{Quattrociocchi2014Opinion}. Their microblogging and decentralized architectures enable participation at unprecedented scale, amplifying civic engagement and the rapid spread of ideas~\cite{DBLP:journals/corr/abs-2211-15988}. Yet these same affordances also fuel harmful dynamics: platforms facilitate misinformation~\cite{DBLP:journals/corr/VicarioVBZSCQ16}, foster ideological echo chambers~\cite{DBLP:journals/pnas/CinelliMGQS21}, and intensify online hostility~\cite{DBLP:conf/cscw/ChengBDL17,DBLP:conf/socinfo/QuattrociocchiE22,DBLP:conf/www/SaveskiRR21}. Such hostile environments often give rise to various forms of antisocial behavior (ASB), including cyberbullying~\cite{DBLP:conf/flairs/OllagnierCV23}, hate speech~\cite{DBLP:journals/snam/OllagnierCV23}, trolling~\cite{doi:10.1177/0267323118760323}, and sexual harassment~\cite{DBLP:conf/acl/ChowdhurySSM19}.

\begin{figure}[htpb]
    \centering
    \includegraphics[width=0.48\textwidth]{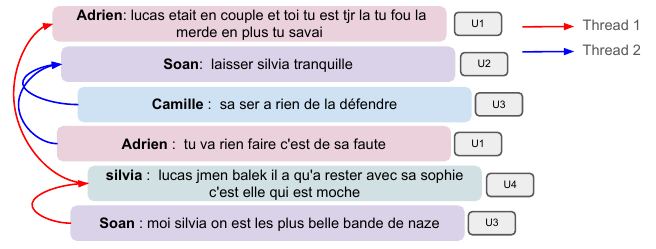}
    \caption{An example of a multi-party interaction, with participants and threads marked, extracted from the \textit{CyberAgressionAdo-Large} dataset~\cite{Ollagnier2024CyberAgressionAdo}. Translation available in Appendix~\ref{sec:trans}.}
    \Description{Example conversation from the CyberAgressionAdo-Large dataset showing a multi-party chat with participants, message threads, and their connections.}
    \label{fig:conv_fr}
\end{figure}

In response, the automatic detection of ASB has become a central research focus in NLP, with numerous shared tasks advancing detection and classification methods~\cite{DBLP:journals/information/AlkomahM22,kumar2025exploring}. Yet most existing work relies on public data from platforms such as Facebook, YouTube, or Instagram and targets relatively narrow subtasks (e.g., binary hate vs. non-hate classification, target identification, or hate-type categorization). At the same time, private messaging apps and chat rooms have become critical arenas for cyberbullying, particularly among adolescents~\cite{alhashmi2023taxonomy}. ASB in these settings remains underexplored, constrained by strict data access restrictions and the complexity of multi-party interactions. Recently released datasets simulating online aggression in private messaging contexts~\cite{DBLP:conf/acl-alw/SprugnoliMTOP18,DBLP:conf/coling/Ollagnier24} offer new opportunities to address this gap. As illustrated in Figure~\ref{fig:conv_fr}, multi-party conversations are non-linear and interleaved, with utterances directed to multiple participants. These dynamics demand computational models that go beyond message-level detection to capture discourse structure, participant roles, and interactional patterns~\cite{ganesh-etal-2023-survey}.

In this paper, we address these limitations by systematically evaluating state-of-the-art representation learning methods---spanning both lexical and graph-based approaches---for ASB detection in multi-party dialogues. Prior work has shown the benefit of combining textual and interactional features; we extend this approach by constructing embeddings that jointly capture semantic content and conversational structure~\cite{DBLP:journals/sncs/CecillonLDL21} and by testing multimodal fusion strategies that integrate both. Experiments are conducted on the \textit{CyberAgressionAdo-Large} dataset~\cite{Ollagnier2024CyberAgressionAdo}, a publicly available corpus of 36 simulated aggressive conversations in French, collected through role-play in educational settings\footnote{\url{https://anonymous.4open.science/r/CyberAgression-Large-C71C}}
. Crucially, to move beyond the narrow subtasks that dominate prior work, we introduce two novel evaluation settings tailored to multi-party contexts: \textit{bullying behavior analysis} (BBA), which classifies the discursive function of messages (bullying vs. non-abusive intent), and \textit{bullying peer group identification} (BPI), which identifies participants' social alignment (e.g., bully vs. victim side). These tasks extend the scope of ASB detection from surface-level classification to pragmatic dimensions of language, capturing intent, stance, and role in dynamic interactions. 
Overall, this paper makes three main contributions: (1) a comprehensive benchmark of unimodal and multimodal representation learning methods for ASB detection in multi-party conversations, (2) the introduction of two pragmatics-oriented evaluation tasks (BBA and BPI) that address group-level and functional dimensions of cyberbullying, and (3) evidence that integrating linguistic and structural features provides a more robust account of the social dynamics underpinning online aggression.

\section{Related Work}
Early work on ASB detection treated messages in isolation, relying on shallow features such as n-grams and sentiment scores. These approaches were soon surpassed by deep learning models (CNNs, RNNs, Transformers) and, more recently, by pretrained language models like BERT and its variants, which now dominate benchmarks on tasks such as target identification and abuse type classification~\cite{DBLP:conf/aiccsa/Alrehili19,DBLP:journals/information/AlkomahM22,kumar2025exploring,ganesh-etal-2023-survey}. Beyond message-level analysis, research has increasingly turned to modeling interactional structure. Graph-based methods capture conversational dynamics by representing participants as nodes and exchanges as edges~\cite{DBLP:journals/sncs/CecillonLDL21,DBLP:conf/coling/Ollagnier24}. Traditional embeddings (node2vec, DeepWalk) exploit topological similarity, while Graph Neural Networks (e.g., GCNs~\cite{DBLP:journals/isci/0004JWLJJH22}, GraphSAGE~\cite{DBLP:journals/tnn/ChenGWWXLLWL22}) enable neighborhood-level aggregation. Signed graph learning further enriches these models by encoding the polarity of interactions (support vs. hostility), which is crucial for distinguishing roles such as aggressor and defender~\cite{DBLP:journals/access/CecillonLDA24}. Early signed embeddings (SNE~\cite{DBLP:conf/ijcai/WangWZJ17}, SIDE~\cite{DBLP:conf/sdm/KimE20}) captured attraction--repulsion patterns but lacked higher-order structure. Newer signed GNNs (SGCN~\cite{DBLP:conf/icdm/Derr0T18}, SiGAT~\cite{wang2020}, SAT~\cite{li2021signed}) address this by integrating polarity-sensitive message passing and balance-aware objectives. Despite these advances, most systems remain either textual or graph-based, with limited integration. Recent multimodal fusion approaches bridge this gap by combining textual embeddings with interactional graphs, improving robustness in scenarios where abusive behavior is distributed across participants or conveyed implicitly~\cite{cheng2020,DBLP:phd/hal/Cecillon24,DBLP:journals/computing/CecillonLD25}.

Progress has also depended on the availability of annotated datasets. Existing resources include ConvAbuse (AI--user dyadic chats)~\cite{DBLP:conf/emnlp/CurryAR21}, the Wikipedia Abuse Corpus (talk page discussions)~\cite{DBLP:conf/lrec/CecillonLDL20}, and the WhatsApp dataset (teen group chats in Italian)~\cite{DBLP:conf/acl-alw/SprugnoliMTOP18}. In French, CyberAgressionAdo V1 and V2 simulate aggressive school-based chats annotated for abuse types, speaker roles, and pragmatic cues~\cite{DBLP:conf/lrec/OllagnierCVB22,DBLP:conf/coling/Ollagnier24}. While valuable, these datasets are small and rarely combine message-level and participant-level annotations, limiting analysis of intent, stance, and role in dynamic interactions.

Building on this foundation, we leverage \textit{CyberAgressionAdo-Large}~\cite{Ollagnier2024CyberAgressionAdo}, the largest publicly available corpus of aggressive multi-party chats in French. This dataset enables detailed study of lexical content, interaction patterns, and participant roles. Unlike prior work focused on narrow subtasks, we propose a unified evaluation across three complementary objectives: abuse detection, bullying behavior analysis, and peer group identification, benchmarking a broad range of textual, graph-based, and fusion-based representation learning methods.

\section{Addressing Online Hate in Conversations}

\subsection{Tasks of Interest}
\label{sec:tasks}

The emergence of datasets simulating online aggression in private instant messaging environments has opened new avenues for studying ASB in multi-party conversations. These resources not only provide previously inaccessible interactional data but also expose the limitations of prevailing approaches. Most prior work reduces ASB detection to binary classification of isolated utterances, overlooking discourse-level context, speaker roles, and pragmatic intent. As a result, systems often fail to distinguish harmful attacks from defensive or conciliatory responses, or to capture the group-based dynamics that underpin bullying.

In conversational settings, the meaning and impact of a message depend heavily on the social dynamics it participates in. Aggressive language may reflect coercion when used by a perpetrator but serve a protective function when voiced by a victim. Likewise, mockery or profanity in peer exchanges may signal bonding rather than abuse. These examples highlight the need for models that go beyond surface-level toxicity detection to capture underlying intent, role asymmetries, and interaction patterns.

Multi-party settings expose limitations of binary, message-isolated toxicity detection. We therefore introduce two evaluation tasks tailored to discourse and roles:

\begin{itemize}
\item[-] \textbf{Bullying Behavior Analysis (BBA)}: Classifies each message as cyberbullying (\texttt{CBB}) or not (\texttt{NO-CBB}), based on predefined discursive roles and their associated intent. The role taxonomy follows prior work~\cite{DBLP:conf/coling/Ollagnier24}. Roles such as \texttt{attack}, \texttt{gaslighting}, and \texttt{instigating/abetting} are labeled \texttt{CBB}, while \texttt{empathy}, \texttt{counterspeech}, \texttt{conflict resolution}, and \texttt{defend} are \texttt{NO-CBB}. This task explicitly separates abusive behavior from reactive or supportive messages, addressing a key limitation of binary hate detection.

\item[-] \textbf{Bullying Peer Group Identification (BPI)}: Assigns each participant to one of four active social roles: \texttt{victim}, \texttt{victim support}, \texttt{bully}, and \texttt{bully support}. The role taxonomy follows prior work~\cite{DBLP:conf/coling/Ollagnier24}. The neutral \texttt{conciliator} role is excluded. Unlike binary classification tasks, BPI captures the peer-group nature of bullying by modeling power asymmetries and affiliations within multi-party interactions.
\end{itemize}

\subsection{Modeling Conversations}
\label{section:modeling}


\begin{figure*}[htpb]
    \centering
  \includegraphics[width=0.6\textwidth]{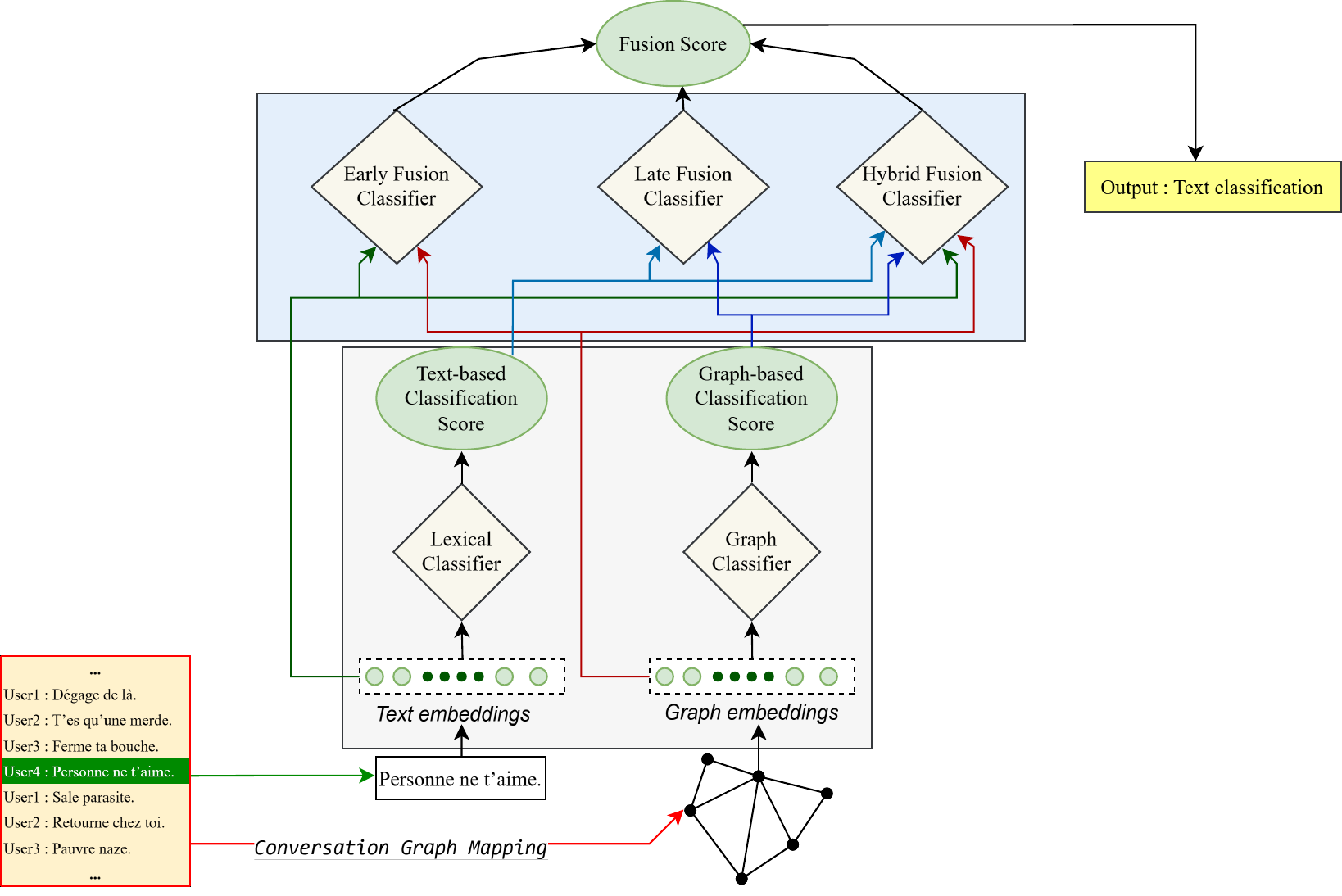}
    \caption{Illustration of the three fusion strategies. Two independent embedding methods produce vector representations used by unimodal classifiers. The fusion phase introduces three classifiers---early, late, and hybrid---each integrating information differently (see main text).}
    \Description{Diagram showing three fusion strategies: early fusion combining embeddings before classification, late fusion merging classifier outputs, and hybrid fusion combining both approaches.}
    \label{fig:pipeline}
\end{figure*}

This section outlines the representation learning methods employed to tackle the three core tasks: abuse detection (ABD), BBA, and BPI. ABD is framed as a binary classification task that determines whether a message contains abusive content, independent of the sender's role or conversational context~\cite{DBLP:conf/coling/Ollagnier24}. To address these tasks, we benchmark a diverse set of text-based, graph-based, and fusion-based models.

\textbf{Lexical embeddings}. To encode the semantic content of individual messages, we use six large language models. These include general-purpose systems---Gemini~\cite{DBLP:journals/corr/abs-2403-20327}, mBERT~\cite{DBLP:journals/corr/abs-1810-04805}, GPT~\cite{gpt4}---and French-specific models: CamemBERT~\cite{DBLP:conf/acl/MartinMSDRCSS20}, FlauBERT~\cite{DBLP:conf/lrec/LeVFSCLACBS20}, and CamemBERTa~\cite{DBLP:journals/corr/abs-2411-08868}. Each model produces contextualized token embeddings, which are aggregated into a global message-level representation. Importantly, messages are encoded independently: no conversational history is aggregated, ensuring that the representation reflects only the local semantics of the message under analysis.


\textbf{Graph embeddings}. To capture conversational structure, we construct directed interaction graphs around each target message. A context window defines the set of surrounding messages, and a sliding window identifies potential recipients. Nodes represent participants, while edges encode message exchanges with attributes for polarity (positive/negative), weight (frequency), and direction (reply flow). We evaluate both node-level and whole-graph embeddings. WD-SGCN emphasizes high-weighted incoming connections, while FGSD and GraphWave capture global structural signatures. WalkLets and Node2Vec rely on random-walk strategies to model topological similarity, though they ignore polarity and weight information. SG2V and its weighted variant WD-SG2V extend this idea by incorporating signed edges, thereby encoding positive and negative interactions. Finally, NGNN merges subgraph representations to capture higher-order structural patterns. Together, these methods differ in how they exploit graph attributes such as directionality, polarity, and weight, offering complementary perspectives on conversational structure.

\textbf{Fusion strategies}. To integrate linguistic and structural signals, we implement three strategies, illustrated in Figure~\ref{fig:pipeline}. Each builds on representations and classifiers trained separately on text and graph modalities~\cite{DBLP:phd/hal/Cecillon24}. In \textit{early fusion}, embeddings from both modalities are concatenated into a unified feature vector for training a new classifier (e.g., SVM), allowing joint learning at the representation level and direct interaction between lexical and structural features. In \textit{late fusion}, prediction scores from unimodal classifiers are fed into a meta-classifier, which combines their outputs and is particularly effective for integrating heterogeneous or noisy signals. Finally, \textit{hybrid fusion} leverages both embedding-level and score-level information, enabling the final classifier to exploit fine-grained features as well as high-level decision patterns, thereby enhancing robustness and adaptability.

\section{Experimental Setup}

This section presents the dataset used to evaluate the three ASB tasks, the graph construction process, the evaluation protocol, and implementation details. Our study addresses two research questions:

\begin{itemize}
    \item[-] \textbf{RQ1}: How do different representation learning modalities (lexical, structural, and fusion-based) perform across the three ASB tasks (ABD, BBA, and BPI)?
    \item[-] \textbf{RQ2}: To what extent do multimodal fusion strategies (early, late, and hybrid) improve the modeling of abusive dynamics and participant roles in multi-party conversations compared to unimodal modalities?
\end{itemize}

Our experiments use the \textit{CyberAgressionAdo-Large} dataset\footnote{Publicly available at: \url{https://anonymous.4open.science/r/CyberAgression-Large-C71C}}, an open-access resource of 36 French multi-party conversations collected in educational settings through structured role-playing games. The dataset simulates realistic scenarios of cyber aggression among adolescents in private chat environments, with participants engaging in behaviors ranging from teasing to overt verbal abuse. Each message is annotated for multiple tasks and enriched with metadata such as discursive roles and conversational context. The corpus covers four sensitive topics and six annotation layers, enabling multi-level analyses. Summary statistics of messages and graphs are reported in Table~\ref{tab:dataset_stats}.

\begin{table*}[ht]
\centering
\caption{Statistics describing the messages (top part) and graphs (bottom part) of our dataset. Symbol ${\pm}$ denotes the standard deviation.}
\begin{tabular}{lccc}
\hline
\textbf{} & \textbf{Average} & \textbf{Minimum} & \textbf{Maximum} \\
\hline
\multicolumn{4}{l}{\textit{Messages}} \\
Number of words per message & 7.22 ${\pm}$ 5.06 & 1 & 94 \\
Number of characters per message & 35.73 ${\pm}$ 25.72 & 1 & 509 \\
Number of messages per conversation & 86.62 ${\pm}$ 50.09 & 42 & 375 \\
\hline
\multicolumn{4}{l}{\textit{Graphs}} \\
Number of vertices per graph & 47.8 ${\pm}$ 20.3 & 2 & 214 \\
Graph density & 0.54 ${\pm}$ 0.15 & 0.12 & 1.00 \\
Number of negative edges per graph & 189.7 ${\pm}$ 38.4 & 3 & 1,800 \\
Number of positive edges per graph & 272.3 ${\pm}$ 45.7 & 5 & 2,500 \\
\hline
\end{tabular}
\label{tab:dataset_stats}
\end{table*}

To model conversational structure, we construct interaction graphs centered on a \textit{target message} (the message to be classified). Graphs are directed and weighted, following the network extraction methodology of~\cite{DBLP:journals/access/CecillonLDA24}. Nodes represent participants, and edges encode co-occurrence of messages within a fixed-size sliding window of 21 surrounding messages. On average, graph sizes are reported in Table~\ref{tab:dataset_stats} ($\approx$48 vertices on average). We refer to that table for the exact range. Polarity is derived from sentiment annotations: messages labeled as positive, negative, or neutral are mapped to signed edges, with neutral treated as positive for experimental purposes. Because polarity is annotation-based rather than predicted, graph structures are inherently language-agnostic. For graph-level models, the full graph is used; for node-level models, only the embedding of the target author node is retained.

For evaluation, the dataset is split into 70\% training and 30\% testing using stratified sampling to preserve class distributions. All models are assessed with 5-fold cross-validation, and performance is reported using the weighted F1-score, with standard deviation across folds to measure stability. Detailed class distributions are provided in Appendix~\ref{appendix:distri}.

As introduced in Section~\ref{section:modeling}, we benchmark six lexical and eight graph-based representation learning methods. We also evaluate fusion strategies that combine top-performing lexical models (\texttt{CamemBERTa}, \texttt{Gemini}, and \texttt{mBERT}) with graph embeddings. All representations are classified using Support Vector Machines (SVM) implemented with cuML and scikit-learn. Experiments were conducted on a machine with an Intel\textregistered Core\texttrademark i7--1065G7 3.5 GHz CPU and an Nvidia A100 GPU. Hyperparameter settings are provided in Appendix~\ref{section:model_setting}.

Finally, the conversation-oriented nature of our dataset and methods limits direct comparison with existing ASB detection tools. Most prior approaches rely on proprietary conversational datasets, which prevents reproducibility~\cite{DBLP:conf/lrec/CecillonLDL20}, while publicly available resources are typically message-level and lack the multi-turn structure required here. Applying such datasets would discard conversational context, leading to incomplete comparisons. Furthermore, most existing tools are trained on English corpora, which hinders their portability to French, as previously noted in the literature~\cite{DBLP:journals/computing/CecillonLD25}.

\subsection{Performance Evaluation}
\label{sec:results}

To address \textbf{RQ1} and \textbf{RQ2}, we evaluate all representation learning methods on the \textit{CyberAgressionAdo-Large} dataset. Table~\ref{tab:appresults} reports the top-performing models, while the full set of results is available in the KIDOS project repository\footnote{\url{https://drive.google.com/drive/folders/1f21Pd1h-VDQAH4hK_bilIMvp4bUves_C?usp=sharing}}.

For \textbf{RQ1}, clear differences emerge between lexical and graph-based encoders. Among lexical models, \texttt{CamemBERT} yields the strongest overall performance (0.697${\pm}$0.01 on ABD, 0.587${\pm}$0.02 on BBA), closely followed by \texttt{Gemini004} with slightly better results on BBA (0.608${\pm}$0.02). By contrast, \texttt{CamemBERTa} and \texttt{Gemini001} consistently lag behind. Yet across all lexical models, performance on BPI remains modest (0.255--0.283), underscoring a fundamental limitation of content-only features: while effective for recognizing abusive language and intent, they fail to capture relational dynamics that define peer-group interactions.

Graph embeddings reveal the opposite pattern. \texttt{WD-SGCN}, which integrates direction, polarity, and edge weights, achieves the highest BPI score among unimodal models (${0.296\pm 0.01}$), highlighting the importance of modeling signed and weighted interactions for role attribution. In contrast, approaches neglecting edge semantics---such as \texttt{Node2Vec}, \texttt{SG2V}, or \texttt{GraphWave}---underperform substantially on BPI ($\leq$0.237). Interestingly, models like \texttt{FGSD} and \texttt{NGNN} yield balanced but unremarkable results, suggesting some sensitivity to structural patterns but insufficient discriminatory power for complex role-based reasoning. Taken together, these results suggest that lexical models specialize in surface-level abuse detection, while graph-based methods are better suited for socially grounded classification, particularly when capturing implicit roles and alignments.

For \textbf{RQ2}, multimodal fusion consistently outperforms unimodal baselines, though task-specific strengths emerge across strategies. 

\textbf{Early fusion}, which concatenates lexical and structural embeddings, proves most effective for BPI. The combination \texttt{mBERT + SG2V} achieves the overall best BPI score (0.301${\pm}$0.01), while \texttt{Gemini004 + GraphWave} (0.299${\pm}$0.02) and \texttt{Gemini004 + FGSD} (0.297${\pm}$0.03) also perform strongly. These results suggest that when peer-group identification requires fine-grained integration of semantic and relational cues, representation-level fusion is the most advantageous.

\textbf{Late fusion}, which ensembles predictions from unimodal classifiers, delivers the highest scores on ABD and BBA. Notably, \texttt{mBERT + WD-SGCN} achieves the top ABD result overall (0.718${\pm}$0.02), while \texttt{Gemini001 + WalkLets} dominates BBA (0.625${\pm}$0.01). Although slightly weaker on BPI $\leq$0.291), late fusion provides the strongest and most balanced improvements on content- and behavior-oriented tasks, indicating that decision-level integration leverages modality-specific strengths without overfitting to noisy cues.
While \texttt{mBERT + WD-SGCN} delivers the \emph{best overall} trade-off across tasks (top ABD and competitive BBA/BPI), the top BPI score is obtained by an early-fusion model (\texttt{mBERT + SG2V}, 0.301). We therefore distinguish per-task bests from the overall best compromise.

\textbf{Hybrid fusion}, which combines embeddings with prediction scores, does not reach the top F1-scores but yields consistently stable performance. For instance, \texttt{mBERT + NGNN} achieves 0.707${\pm}$0.02 on ABD, 0.613${\pm}$0.01 on BBA, and 0.286${\pm}$0.00 on BPI. The relatively low variance across folds suggests stronger robustness and adaptability---an important property in real-world applications where conversational dynamics and abuse strategies are highly variable.

Overall, lexical models perform best on content-oriented tasks (ABD, BBA), while graph-based encoders show relative advantages on socially grounded reasoning, particularly in BPI, where modeling interaction structure is essential. Yet absolute performance on BPI remains modest across models, reflecting the difficulty of this four-class, highly imbalanced task and pointing to the need for richer approaches to capture peer-group dynamics. Fusion strategies consistently outperform unimodal baselines, with early fusion proving most effective for relational inference, late fusion excelling at content-driven detection, and hybrid fusion offering robust generalization. Taken together, these results underscore the complementary strengths of lexical and structural cues and highlight the necessity of their integration to model the nuanced and evolving dynamics of abusive behavior in multi-party online conversations.

\begin{table*}[ht]
     \centering
\caption{Average weighted F-scores (mean $\pm$ std. from 5--fold cross-validation) with SVM classifier. Bold marks the best score per task; The row of best overall model is highlighted. Results from other classifiers are in Appendix~\ref{section:model_setting}.}
     \begin{tabular}{lccccc}
     \toprule 
     \textbf{Embedding Type} & \textbf{Dimension} & \textbf{ABD} & \textbf{BBA} & \textbf{BPI} \\
     \midrule
    \textbf{\textit{Lexical embedding}} & & & & \\
    \texttt{Gemini004} & 1024 & 0.676${\pm}$0.02 & 0.608${\pm}$0.02 & 0.283${\pm}$0.02 \\
    \texttt{CamemBERT} & 768 & 0.697${\pm}$0.01 & 0.587${\pm}$0.02 & 0.279${\pm}$0.02 \\
    \texttt{CamemBERTa} & 768 & 0.660${\pm}$0.01 & 0.590${\pm}$0.01 & 0.255${\pm}$0.01 \\
    \texttt{Gemini001} & 768 & 0.656${\pm}$0.01 & 0.572${\pm}$0.02 & 0.267${\pm}$0.02 \\
    \midrule
    \textbf{\textit{Graph embedding}} & & & & \\
    \texttt{WD-SGCN} & 128 & 0.566${\pm}$0.02 & 0.479${\pm}$0.03 & 0.296${\pm}$0.01 \\
    \texttt{FGSD} & 200 & 0.529${\pm}$0.00 & 0.533${\pm}$0.00 & 0.236${\pm}$0.02 \\
    \texttt{WalkLets} & 32 & 0.488${\pm}$0.00 & 0.516${\pm}$0.02 & 0.237${\pm}$0.00 \\
    \texttt{WD-SG2V} & 128 & 0.507${\pm}$0.00 & 0.475${\pm}$0.02 & 0.211${\pm}$0.00 \\
    \texttt{NGNN} & 64 & 0.483${\pm}$0.01 & 0.529${\pm}$0.01 & 0.178${\pm}$0.01 \\
    \texttt{SG2V} & 128 & 0.481${\pm}$0.00 & 0.444${\pm}$0.01 & 0.231${\pm}$0.00 \\
    \texttt{Node2Vec} & 128 & 0.467${\pm}$0.00 & 0.465${\pm}$0.02 & 0.224${\pm}$0.01 \\
    \texttt{GraphWave} & 200 & 0.473${\pm}$0.00 & 0.438${\pm}$0.00 & 0.237${\pm}$0.02 \\
    \midrule
    \textbf{\textit{Fusion embedding}} & & & & \\

    \textbf{Early Fusion} & & & & \\
    \texttt{Gemini004 + GraphWave} & 1152 & 0.684${\pm}$0.02 & 0.616${\pm}$0.01 & 0.299${\pm}$0.02 \\
    \texttt{mBERT + SG2V} & 896 & 0.703${\pm}$0.01 & 0.590${\pm}$0.02 & \textbf{0.301${\pm}$0.01} \\
    \texttt{Gemini004 + FGSD} & 776 & 0.674${\pm}$0.01 & 0.617${\pm}$0.01 & 0.297${\pm}$0.03 \\

    \textbf{Late Fusion} & & & & \\
    \texttt{Gemini004 + WD-SG2V} & 1152 & 0.715${\pm}$0.02  & 0.616${\pm}$0.03 & 0.284${\pm}$0.00 \\
    \texttt{Gemini001 + WalkLets} & 1032 & 0.697${\pm}$0.02  & \textbf{0.625${\pm}$0.01} & 0.291${\pm}$0.00 \\
    \texttt{mBERT + WD-SGCN} & 776 & \mycc{\textbf{0.718${\pm}$0.02}} & \mycc 0.606${\pm}$0.01 & \mycc 0.286${\pm}$0.00 \\

    \textbf{Hybrid Fusion} & & & & \\
    \texttt{mBERT + NGNN} & 1032 & 0.707${\pm}$0.02  & 0.613${\pm}$0.01 & 0.286${\pm}$0.00 \\
    \texttt{Gemini004 + Node2Vec} & 896 & 0.696${\pm}$0.02  & 0.617${\pm}$0.03 & 0.277${\pm}$0.00 \\
    \texttt{CamemBERTa + WD-SGCN} & 776 & 0.698${\pm}$0.01  & 0.604${\pm}$0.02 & 0.267${\pm}$0.00 \\
    \bottomrule
    \end{tabular}
    \label{tab:appresults}
\end{table*}

\subsection{Error Analysis}

To deepen our understanding of how different representation learning strategies handle abusive dynamics, we conducted a systematic error analysis across the three classification tasks. Our goal was to evaluate not only overall accuracy but also each model's capacity to capture participant roles, pragmatic cues, and evolving discourse structures in multi-party conversations. \\
All misclassification counts in Table~\ref{tab:misclassification_breakdown} are aggregated over the five test splits in our cross-validation. For each model, we collect the misclassified instances from the test set of every split and sum them to obtain the total per class. The reported values therefore reflect cumulative errors across all folds, not results from a single split.

\begin{table*}[!htbp]
\centering
\caption{Misclassified instance counts and percentages per label across tasks and models. Models are grouped by representation type: Text, Graph, and Fusion.}
\label{tab:misclassification_breakdown}

\resizebox{\textwidth}{!}{%
\begin{tabular}{ll|ccc|ccc|cccc}
\toprule
\textbf{Task} & \textbf{Label} & \multicolumn{3}{c|}{\textbf{Text Models}} & \multicolumn{3}{c|}{\textbf{Graph Models}} & \multicolumn{4}{c}{\textbf{Fusion Models}} \\
\cmidrule(r){3-5} \cmidrule(r){6-8} \cmidrule(r){9-12}
& & Gemini004 & CamemBERT & mBERT & GraphWave & SG2V & FGSD & Gemini004 + GW & CamemBERT + SG2V & mBERT + SG2V & Gemini004 + FGSD \\
\midrule
\multirow{2}{*}{\textbf{ABD}} 
& Non-abusive  & 47.45\% (298) & 47.77\% (300) & 28.57\% (104) & 40.13\% (252) & 40.76\% (256) & 49.68\% (312) & 47.45\% (298) & 45.38\% (285) & 45.38\% (285) & 49.68\% (312) \\
& Abusive      & 22.45\% (229) & 18.53\% (189) & 34.60\% (91)  & 37.65\% (384) & 36.47\% (372) & 21.08\% (215) & 20.98\% (214) & 19.12\% (195) & 19.12\% (195) & 21.08\% (215) \\
\midrule
\multirow{2}{*}{\textbf{BBA}} 
& Non-bullying & 37.14\% (361) & 39.40\% (383) & 30.32\% (67)  & 42.80\% (416) & 42.39\% (412) & 37.45\% (364) & 35.49\% (345) & 37.96\% (369) & 37.96\% (369) & 37.45\% (364) \\
& Bullying     & 42.75\% (289) & 44.67\% (302) & 33.70\% (92)  & 32.54\% (220) & 31.95\% (216) & 40.09\% (271) & 43.05\% (291) & 45.86\% (310) & 45.86\% (310) & 40.09\% (271) \\
\midrule
\multirow{4}{*}{\textbf{BPI}} 
& Victim         & 25.70\% (165) & 25.46\% (180) & 25.46\% (180) & 25.79\% (164) & 25.63\% (162) & 25.83\% (202) & 25.70\% (165) & 29.86\% (209) & 29.86\% (209) & 28.63\% (201) \\
& Victim Support & 32.55\% (209) & 29.84\% (211) & 29.84\% (211) & 32.08\% (204) & 31.96\% (202) & 30.18\% (236) & 32.55\% (209) & 35.00\% (245) & 35.00\% (245) & 33.76\% (237) \\
& Bully          & 21.03\% (135) & 22.49\% (159) & 22.49\% (159) & 21.23\% (135) & 21.36\% (135) & 22.12\% (173) & 21.03\% (135) & 18.00\% (126) & 18.00\% (126) & 19.09\% (134) \\
& Bully Support  & 20.72\% (133) & 22.21\% (157) & 22.21\% (157) & 20.91\% (133) & 21.04\% (133) & 21.87\% (171) & 20.72\% (133) & 17.14\% (120) & 17.14\% (120) & 18.52\% (130) \\
\bottomrule
\end{tabular}
}
\end{table*}

\subsubsection*{RQ1 : Modality-wise performance across ABD, BBA, and BPI}

\paragraph{\textbf{ABD}}
Table~\ref{tab:misclassification_breakdown} shows that text-based models misclassify a substantial portion of \emph{Abusive} messages (e.g., Gemini004: 22.45\% / 229). In contrast, graph-based models yield higher error rates for both \emph{Abusive} messages (up to 37.65\% / 384) and \emph{Non-abusive} cases (FGSD: 49.68\% / 312), indicating a systematic over-detection bias. 

Misclassified instances often involve implicit aggression, sarcasm, or connotative tone---features that are not structurally salient. For example, the message \emph{``elle est bonne la meuf''} (EN: \emph{``That girl is hot''}), labeled as abusive for its objectifying tone, was missed by GraphWave but correctly identified by Gemini004, highlighting that lexical cues remain critical for capturing implicit forms of abuse.

\paragraph{\textbf{BBA}}
Both text and fusion models accumulate high error rates on the \emph{Bullying} label (43--46\% misclassified, Table~\ref{tab:misclassification_breakdown}), showing persistent difficulty in distinguishing bullying from non-bullying when cues are indirect. Graph-based models distribute errors more evenly but still misclassify about one third of both classes. For instance, the message \emph{``c'est toi qui commence a chercher les gens''} (EN: \emph{You're the one who starts provoking people}) was misclassified as \textsc{No-CBB} by GraphWave, while Gemini004 correctly inferred the accusatory stance, illustrating the lexical vs.\ structural contrast.

\paragraph{\textbf{BPI}}
Errors are not uniform across roles. \emph{Victim Support} is systematically the most misclassified (CamemBERT: 24.65\% / 211, GraphWave: 23.83\% / 204, FGSD: 27.57\% / 236). In contrast, explicit aggressor roles (\emph{Bully}, \emph{Bully Support}) show lower error shares ($\approx$15--22\%), suggesting they are easier to detect. Example: the message \emph{``tu fais le mec intelligent pour impressionner les autres''} (EN: \emph{You're pretending to be smart to impress others}) was correctly flagged as aggressive by Gemini004 but missed by GraphWave, confirming that subtle antagonism requires lexical sensitivity.

 Across modalities, the hardest cases are implicit abuse (ABD, BBA) and ambiguous supportive roles (BPI). Graph encoders over-predict abuse, while text encoders miss subtler aggression.

\subsubsection*{RQ2 : Impact of fusion strategies on modeling abuse and roles}

\paragraph{\textbf{ABD}} Fusion reduces misclassifications on the \emph{Abusive} class (down to 19.12\% / 195) compared to 22--38\% with single modalities. However, this comes at the cost of slightly more false positives on \emph{Non-abusive} samples ($\approx$45\% vs.\ 28\% for mBERT alone).

\paragraph{\textbf{BBA}} Fusion decreases errors on \emph{Non-bullying} (35.49\% / 345 vs.\ 37--42\% with single models), but errors on \emph{Bullying} remain high ($\approx$40--46\%), indicating that the main source of confusion persists.

\paragraph{\textbf{BPI}} Fusion particularly reduces errors on aggressor roles. \emph{Bully} misclassifications drop from 22.49\% (159, text) to 18.00\% (126, fusion), and \emph{Bully Support} from 22.21\% (157) to 17.14\% (120). At the same time, \emph{Victim} and \emph{Victim Support} errors increase, reflecting a redistribution of mistakes rather than a global reduction.

\paragraph{\textbf{Summary.}} 
Fusion strategies shift rather than eliminate errors. They reduce false negatives for aggressor roles (ABD \emph{Abusive}, BPI \emph{Bully}) but increase confusion for neutral or supportive roles. This pattern highlights complementary error profiles between text and graph modalities and underscores the need for adaptive fusion mechanisms sensitive to role dynamics and pragmatic subtleties.

\section{Conclusion and Future directions}
This paper presents representation learning methods for detecting hate speech in multi-party dialogues and introduces two new evaluation tasks: Bullying Behavior Analysis (BBA) and Bullying Peer Group Identification (BPI). These tasks provide a more granular understanding of how aggressive behaviors emerge and evolve within multi-participant conversations. We benchmark a broad spectrum of lexical and graph-based embedding models and assess fusion strategies that combine linguistic content with structural interaction patterns. Results show that multimodal approaches consistently outperform unimodal baselines---particularly on socially complex tasks like BPI, where understanding implicit roles and interactional dynamics is crucial. Lexical models excel in tasks driven by explicit textual cues (ABD, BBA), while graph-based models are better suited to relational inference. Fusion strategies, especially early fusion, capitalize on the complementary strengths of both modalities, yielding the most robust and context-aware representations for abuse detection.

In future work, more integrated representation learning could be explored by jointly encoding lexical and structural features within a unified model, rather than learning them separately. Additionally, while our experiments rely on structured role-play data for ethical and practical reasons, validating models on real-world conversations across different platforms and demographics will be critical for assessing generalizability. Incorporating richer context, such as temporal dynamics, user-level features, and conversational history, may further enhance the detection of subtle or evolving aggression patterns.

\section{Limitations}
While our evaluation is based on \textit{simulated} cyber aggression data generated through controlled role-play, this method has been shown to produce more naturalistic language than interviews, questionnaires, human--machine interactions, or reconstructed threads~\cite{tran2006naturalized}. However, as with any simulated resource, there may be differences from spontaneous online conversations, particularly in tone, fluency, or conversational dynamics. Moreover, cultural and linguistic features specific to French-speaking adolescents may shape language use and social behaviors, potentially limiting the generalizability of our findings. Future work should validate the models on naturally occurring data across more diverse linguistic and cultural settings to ensure cross-context robustness.



Furthermore, while our tasks span both binary (e.g., ABD, BBA) and multi-class (e.g., BPI) classification, mapping complex social behaviors to fixed labels can obscure key contextual nuances. For instance, teasing among peers or reactive aggression may be misclassified without considering interpersonal dynamics or conversation history~\cite{cowie2014understanding}. Additionally, our use of undersampling to mitigate class imbalance ensures balanced training but may distort the natural prevalence of abusive behaviors, which are typically rare in real-world settings. Future work should explore alternative approaches---such as oversampling, cost-sensitive learning, or data augmentation---to enhance model robustness and better reflect realistic distributions.

An additional limitation lies in the current binary design of signed conversational graphs, which model only positive and negative interactions between participants. While effective in capturing antagonistic or affiliative dynamics, this dichotomy overlooks the prevalence and pragmatic significance of neutral exchanges---such as factual statements, inquiries, or disengaged responses---that play a crucial role in structuring online conversations~\cite{DBLP:conf/flairs/OllagnierCV23}. Our current experiments map neutral messages to positive edges for simplification, which may conflate passive or ambiguous behavior with active support. Future work should incorporate a third edge type to explicitly represent neutral interactions, thereby enhancing the expressiveness of conversational graphs and enabling a more nuanced modeling of participant roles, pragmatic stance, and socio-discursive functions in cyber aggression.

\section{Ethical Considerations}

Despite its benefits, role-play data may still expose both participants and annotators to emotionally charged or distressing situations. To mitigate these risks, we selected this dataset specifically because it was created following a strict experimental protocol for participants and adhered to established ethical and procedural standards for annotation. During data collection, all student participants under the age of 18 provided informed parental consent, and each school's administration and ethics board approved the study. Psychological support was available throughout, and students participated in dedicated debriefing sessions focused on cyberbullying awareness and digital safety~\cite{DBLP:conf/lrec/OllagnierCVB22}. Importantly, no student was asked to play the role of a victim, and all sessions were conducted under the supervision of trained researchers ready to intervene if necessary.

The annotation process was similarly designed to minimize emotional burden and ensure labeling consistency, following best practices recommended in~\cite{vidgen-etal-2019-challenges,DBLP:conf/emnlp/KirkBVD22}. Annotators followed a structured training protocol that included reviewing detailed task guidelines and public definitions of antisocial behaviors, studying annotated examples---including edge cases such as implicit abuse or sarcasm---participating in trial labeling sessions with feedback, and engaging in regular calibration meetings to resolve disagreements and align interpretations. Institutional oversight, content warnings, and limited exposure protocols were also implemented to safeguard annotator well-being throughout the process; further details can be found in~\cite{Ollagnier2024CyberAgressionAdo}.

From a modeling perspective, inferring behavioral roles and intentions raises ethical and epistemological concerns, as assigned labels often rely on subjective interpretation and lack objective ground truth---even though the conversations in this dataset were collected through scenario-based role-play. Supporting subjectivity in NLP annotation is essential, as it allows models to reflect and explain the diverse perspectives of real individuals. In this context, the dataset provides annotations from three independent annotators, opening the door to perspectivist approaches~\cite{DBLP:conf/aaai/CabitzaCB23} that embrace disagreement and model a plurality of viewpoints.

Importantly, our goal is to benchmark representation learning strategies, not to propose automated moderation tools. Any real-world deployment must adopt a human-in-the-loop approach, remain transparent about its assumptions, and continuously validate model outputs against sociocultural norms. ASB detection should support---not replace---educators, moderators, or mental health professionals. Finally, care must be taken not to stigmatize youth behavior or overinterpret linguistic features without attention to context, intent, and interpersonal dynamics.

\printbibliography
\appendix
\section{Conversation Translation}
\label{sec:trans}

\begin{figure}[htpb]
    \centering
    \includegraphics[width=0.48\textwidth]{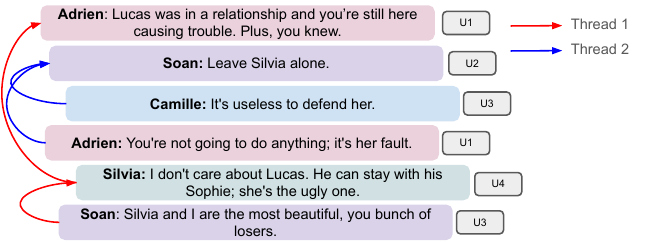}
    \caption{The translated version of Figure~\ref{fig:conv_fr}.}
    \Description{English translation of the original conversation figure: a multi-party chat with labeled participants and threaded replies, showing interactions and role annotations.}
    \label{fig:conv_en}
\end{figure}

\section{Dataset Distribution}
\label{appendix:distri}

For each dataset, we randomly sampled 70\% of the exchanged messages for training and 30\% for testing. To maintain consistency and readability, each binary classification scheme is represented using positive and negative classes. For the ABD task, the positive class consists of messages that perpetrate abusive behaviors, while the negative class includes non-abusive messages. For the BBA task, messages exhibiting cyberbullying behaviors are classified as negative, whereas messages that do not contain cyberbullying behaviors belong to the positive class. For the BPI task, each label represents a different role that participants play in cyberbullying interactions, as described in~\cite{DBLP:conf/lrec/OllagnierCVB22}.
\FloatBarrier
\begin{table}[t]
    \centering
    \caption{Distribution of data splits across tasks.}
    \label{tab:distri_transposed}
    \small
    \setlength{\tabcolsep}{3pt}
    \begin{tabular}{ll*{10}{c}}
        \toprule
        \textbf{Task} & \textbf{Label} & \multicolumn{2}{c}{Split 1} & \multicolumn{2}{c}{Split 2} & \multicolumn{2}{c}{Split 3} & \multicolumn{2}{c}{Split 4} & \multicolumn{2}{c}{Split 5} \\
        \cmidrule(lr){3-4} \cmidrule(lr){5-6} \cmidrule(lr){7-8} \cmidrule(lr){9-10} \cmidrule(lr){11-12}
        & & Tr & Te & Tr & Te & Tr & Te & Tr & Te & Tr & Te \\
        \midrule
        \multirow{2}{*}{ABD} & Pos & 445 & 204 & 445 & 204 & 444 & 204 & 445 & 204 & 445 & 204 \\
                             & Neg & 213 & 125 & 213 & 125 & 214 & 126 & 214 & 126 & 214 & 126 \\
        \midrule
        \multirow{2}{*}{BBA} & Pos & 372 & 194 & 373 & 194 & 373 & 195 & 373 & 195 & 373 & 194 \\
                             & Neg & 286 & 135 & 285 & 135 & 285 & 135 & 286 & 135 & 286 & 136 \\
        \midrule
        \multirow{5}{*}{BPI} & Victim         & 121 & 44 & 122 & 44 & 122 & 44 & 122 & 44 & 121 & 44 \\
                             & Victim Supp.   & 170 & 78 & 170 & 79 & 170 & 79 & 171 & 78 & 171 & 78 \\
                             & Bully          & 154 & 62 & 154 & 62 & 154 & 62 & 153 & 63 & 153 & 63 \\
                             & Bully Supp.    & 167 & 96 & 167 & 96 & 167 & 96 & 167 & 96 & 168 & 96 \\
        \bottomrule
    \end{tabular}
\end{table}
\FloatBarrier

\section{Model Detailed Evaluation}
\label{section:model_setting}
All models are trained using the same pipeline, with SVM hyperparameter configurations detailed in Table~\ref{tab:svm_config}. A grid search is conducted over predefined parameter values, and each configuration is evaluated using 5-fold cross-validation. The optimal hyperparameters are selected based on the highest average weighted F-score. Detailed configurations and results are provided in the KIDOS project folder\footnote{\url{https://drive.google.com/drive/folders/1f21Pd1h-VDQAH4hK_bilIMvp4bUves_C?usp=sharing}}.

\FloatBarrier
\begin{table}[ht!]
  \centering
  \caption{SVM hyperparameter grid. Bold values mark the best-performing parameters.}
  \label{tab:svm_config}
  \small
  \setlength{\tabcolsep}{4pt}
  \begin{tabular}{@{}ll@{}}
    \toprule
    \textbf{Parameter} & \textbf{Values} \\
    \midrule
    Kernel         & \{rbf, sigmoid, \textbf{poly}, linear\} \\
    C              & \{0.01, \textbf{0.1}, 1, 10, 100\} \\
    Gamma          & \{scale, auto, 0.001, \textbf{0.01}, 0.1, 1, 10\} \\
    Degree         & \{2, \textbf{3}, 4, 5\} \\
    Max Iterations & \{100, 200, \textbf{500}\} \\
    \bottomrule
  \end{tabular}
\end{table}
\FloatBarrier

\end{document}